\documentclass[sigconf]{acmart}

\usepackage{makecell}
\usepackage{xcolor}
\usepackage{hyperref}
\usepackage{multirow}
\usepackage{graphicx}
\usepackage{subcaption}

\AtBeginDocument{%
  }

\copyrightyear{2026}
\acmYear{2026}
\setcopyright{cc}
\setcctype{by}
\acmConference[CIKM '26]{Proceedings of the 35th ACM International Conference on Information and Knowledge Management}{November 07--11, 2026}{Rome, Italy}
\acmBooktitle{Proceedings of the 35th ACM International Conference on Information and Knowledge Management (CIKM '26), November 07--11, 2026, Rome, Italy}
\acmDOI{10.1145/3799682.3840775}
\acmISBN{979-8-4007-2539-5/2026/11}

\begin{document}

\title[GT-PSSM for Multivariate Time Series Anomaly Detection]{GT-PSSM: Unified Probabilistic Framework for Stochastic Dynamics Modeling and Dependency Learning in Multivariate Time Series Anomaly Detection}

\author{Wonmo Koo}
\email{joseph92@kaist.ac.kr}
\orcid{0000-0003-0028-2282}
\affiliation{%
  \institution{KAIST}
  \city{Daejeon}
  \country{Republic of Korea}
}
\affiliation{%
  \institution{Pusan National University}
  \city{Busan}
  \country{Republic of Korea}
}
\author{Jaeyeong Lee}
\email{dlwodud116@kaist.ac.kr}
\orcid{0009-0000-0929-2650}
\affiliation{%
  \institution{KAIST}
  \city{Daejeon}
  \country{Republic of Korea}
}

\author{Taeseong Yoon}
\email{bigstar0423@kaist.ac.kr}
\orcid{0009-0007-9810-8827}
\affiliation{%
  \institution{KAIST}
  \city{Daejeon}
  \country{Republic of Korea}
}

\author{Heeyoung Kim}
\authornote{Corresponding author}
\email{heeyoungkim@kaist.ac.kr}
\orcid{0000-0001-6415-9887}
\affiliation{%
  \institution{KAIST}
  \city{Daejeon}
  \country{Republic of Korea}
}

\begin{abstract}
Multivariate time series anomaly detection (MTAD) is crucial for ensuring the safe and reliable operation of complex systems. Many existing methods learn normal patterns by training reconstruction or forecasting models on predominantly normal data. However, a large portion of these approaches rely on deterministic models and their associated point-wise output errors for anomaly scoring. Since real-world multivariate time series are inherently stochastic due to measurement noise and intrinsic system randomness, purely error-based scores can be unreliable, as large errors may arise from benign fluctuations rather than true anomalies. Probabilistic approaches address this limitation by quantifying uncertainty in model outputs. In particular, probabilistic state-space models (PSSMs) provide a principled framework by modeling stochastic system dynamics through latent state transitions and measurement noise via emission models. 
Despite this advantage, existing PSSM-based MTAD methods often struggle to capture long-range temporal dependencies and inter-variable dependencies, as they typically rely on noise-sensitive recurrent architectures and lack explicit cross-variable structure modeling. To address these limitations, we propose Graph-Transformer-Enhanced Probabilistic State-Space Model (GT-PSSM), a novel PSSM-based MTAD method that tightly integrates PSSM-based probabilistic modeling of stochastic dynamics with Graph Transformer-based learning of temporal and inter-variable dependencies. Specifically, GT-PSSM leverages temporal multi-head self-attention to capture long-range temporal dependencies and graph convolution over a learnable graph to explicitly model inter-variable dependencies, while retaining principled stochastic modeling of system dynamics and measurement noise. By jointly modeling stochasticity, long-range temporal dependence, and variable interactions within a unified probabilistic framework, GT-PSSM enables more robust anomaly detection.  Extensive experiments on real-world benchmark datasets demonstrate that GT-PSSM achieves state-of-the-art performance.
\end{abstract}




\maketitle
\section{Introduction}
Multivariate time series are ubiquitous in modern real-world systems, such as internet service infrastructures \cite{su2019robust, abdulaal2021practical}, cyber-physical systems \cite{ahmed2017wadi, mathur2016swat}, smart manufacturing systems \cite{kim2023contextual}, environmental monitoring networks \cite{koo2024deep}, and traffic management systems \cite{jung2024spatially}. These data are often high-dimensional and exhibit complex temporal dynamics and inter-variable dependencies, and are inherently stochastic due to measurement noise and intrinsic randomness in system dynamics. In such systems, abnormal events---such as equipment failures and cyber-attacks---often manifest as anomalous patterns in observed time series. Detecting these anomalies in a timely and reliable manner is crucial for ensuring system safety, stability, and efficiency, as well as for reducing operational costs and preventing catastrophic failures. However, the high dimensionality, complex temporal dynamics and dependencies among variables, and inherent stochasticity of multivariate time series make anomaly detection a particularly challenging task. 

In practice, anomalies are rare and labeled abnormal data are often scarce. Consequently, most multivariate time series anomaly detection (MTAD) methods learn representations of normal behavior from predominantly normal data via reconstruction, forecasting, or hybrid (combining both) objectives \cite{zamanzadeh2024deep}. Under this framework, many existing approaches \cite{deng2021graph, chen2021learning, zheng2023correlation, xu2021anomaly, tuli2022tranad, lai2023nominality, ding2023mst} adopt deterministic time series models and identify anomalies based on point-wise output errors (i.e., reconstruction or forecasting errors), implicitly assuming that larger errors indicate abnormal behavior. However, for inherently stochastic time series, normal random fluctuations can also yield large prediction or reconstruction errors. As a result, purely error-based anomaly scores may be overly sensitive and unreliable, making it difficult to distinguish true anomalies from benign variability \cite{feng2021time, li2020anomaly, feng2024sensitivehue}. 

Probabilistic approaches address this limitation by explicitly quantifying uncertainty in model outputs, allowing anomaly scores to account not only for the magnitude of deviations but also for the expected variability of observations. This yields a more robust mechanism for anomaly detection under stochastic dynamics. 
Accordingly, several MTAD methods  \cite{feng2021time, li2023prototype, feng2024sensitivehue, chen2022deep, feng2024spatial, zhao2020multivariate} employ probabilistic reconstruction or forecasting models---where, in hybrid architectures, either or both components may be probabilistic---and typically score anomalies using the negative log-likelihood (NLL) (or its upper-bound surrogate) of observations under the predictive distribution. By accounting for uncertainty in model outputs, likelihood-based scoring becomes less sensitive to benign stochastic fluctuations and better distinguishes abnormal behavior from expected randomness. 

As a principled probabilistic approach, a line of work \cite{su2019robust, li2020anomaly, dai2021sdfvae, feng2021time, li2022learning, chen2022deep, 9796836} leverages probabilistic state-space models (PSSMs), which represent multivariate time series via a sequence of stochastic latent states  (i.e., per-time-step latent variables). In PSSMs, the temporal evolution of latent states is governed by a state-transition model, enabling explicit modeling of temporal dependencies in stochastic system dynamics, while an emission model relates latent states to observations and accounts for measurement noise. The predictive distribution is then obtained by inferring latent states using either an inference network trained under a variational autoencoder (VAE) \cite{kingma2013auto} framework \cite{li2022learning, 9796836} or Bayesian filtering algorithms~\cite{feng2021time}.

However, existing PSSM-based MTAD methods may still struggle to capture complex temporal and inter-variable dependencies. In particular, latent stochastic states alone may be insufficient to effectively preserve historical information necessary for modeling long-range temporal dependencies \cite{su2019robust, tang2021probabilistic}. 
To address this limitation, existing methods typically condition transition and/or emission distributions on deterministic hidden states of recurrent neural networks (RNNs) to summarize historical information. However, RNNs often fail to preserve reliable long-range information in highly stochastic time series \cite{tuli2022tranad}, limiting their effectiveness in this role. Furthermore, explicit modeling of inter-variable dependencies is often absent, limiting performance in complex multivariate settings.
%However, existing PSSM-based MTAD methods may still struggle with complex temporal and inter-variable dependencies. In particular, they typically condition transition and/or emission distributions on deterministic hidden states of recurrent neural networks (RNNs) to summarize history, but RNNs often fail to preserve reliable long-range information in highly stochastic series \cite{tuli2022tranad}, and explicit modeling of inter-variable dependencies is frequently lacking, limiting performance in complex multivariate settings. 

In this paper, we propose a novel PSSM-based MTAD method, called
\underline{G}raph-\underline{T}ransformer-Enhanced \underline{P}robabilistic \underline{S}tate-\underline{S}pace \underline{M}odel (GT-PSSM).
GT-PSSM performs probabilistic reconstruction for anomaly detection, preserving the strengths of prior PSSM-based methods in modeling stochasticity while overcoming their limitations in capturing long-range temporal and inter-variable dependencies. Specifically, it unifies PSSM-based probabilistic modeling of stochastic dynamics with  Graph Transformer-based learning of temporal and inter-variable dependencies.  At each time step, GT-PSSM conditions both the state-transition and emission distributions on a representation of past observations encoded by a Graph Transformer composed of a temporal multi-head self-attention (MHSA) layer and a graph convolution (GConv) layer. Unlike RNNs, the MHSA layer can effectively capture both short- and long-range temporal dependencies through direct interactions across time steps. Meanwhile, the GConv layer explicitly models inter-variable dependencies through a learnable graph, where nodes represent variables and weighted edges encode their relationships.

In our architecture, the Graph Transformer primarily captures dependency structures through the emission means. Accordingly, we adopt $\beta$-NLL \cite{seitzer2022pitfalls} to stabilize the learning of the emission means while preserving uncertainty modeling. Furthermore, we encourage the dependency structure encoded in the variable-wise emission means to align with the variable interactions learned by the Graph Transformer, thereby further enhancing its ability to capture complex inter-variable dependencies. Through these modeling choices, GT-PSSM tightly integrates the Graph Transformer into the PSSM framework. GT-PSSM incorporates an amortized inference network to infer stochastic latent states for probabilistic reconstruction. The main contributions of this work are summarized as follows:
\begin{itemize}
    \item We propose GT-PSSM, a probabilistic reconstruction-based MTAD method that jointly captures stochasticity, temporal dependencies, and inter-variable dependencies in complex multivariate time series.
    \item GT-PSSM provides a principled PSSM framework that explicitly models both measurement noise and latent stochastic system dynamics.
    \item To overcome the limitations of prior PSSM-based methods, GT-PSSM parameterizes both the state-transition and emission distributions though Graph Transformer representations, integrating temporal self-attention for long-range temporal dependency modeling with graph convolution over a learnable weighted graph for explicit inter-variable dependency modeling, thereby tightly coupling Graph Transformer-based dependency modeling with probabilistic state-space modeling.
    \item Extensive experiments on real-world benchmark datasets demonstrate that GT-PSSM achieves state-of-the-art performance for MTAD.
\end{itemize}

\section{Related Work}
Existing MTAD methods typically learn normal patterns by training reconstruction models \cite{li2019mad, audibert2020usad, xu2021anomaly, tuli2022tranad, lai2023nominality, feng2024sensitivehue, li2023prototype,  li2022learning, dai2024sarad, shimillas2026low} or forecasting models \cite{deng2021graph, chen2021learning, zheng2023correlation,  kim2023contextual2, liu2024multivariate, febrinanto2025entropy, lin2023hybridad, feng2021time, lee2026knowledge} on predominantly normal data. In addition, hybrid models \cite{han2022learning, ding2023mst, chen2022deep, zhao2020multivariate, feng2024spatial} that jointly leverage reconstruction and forecasting have been proposed. These methods can be further categorized into deterministic  \cite{deng2021graph, chen2021learning, zheng2023correlation, xu2021anomaly, tuli2022tranad, lai2023nominality, ding2023mst} and probabilistic approaches \cite{feng2021time, li2023prototype, feng2024sensitivehue, chen2022deep, feng2024spatial, zhao2020multivariate}, depending on the anomaly scoring mechanism. Specifically, deterministic approaches compute anomaly scores based on point reconstruction or forecasting errors et each time step, whereas probabilistic approaches use the NLL (or an upper-bound surrogate) of observations under the predictive distribution. In hybrid settings, we categorize a method as probabilistic if at least one component employs likelihood-based scoring; otherwise, it is regarded as deterministic.

Deterministic approaches have mainly focused on modeling temporal and inter-variable dependencies, as anomalous values often exhibit dependency patterns that deviate from those observed under normal operating conditions. Early approaches primarily employed RNNs \cite{malhotra2015long, hundman2018detecting, li2019mad, wu2020developing} and convolutional neural networks (CNNs) \cite{munir2018deepant, thill2021temporal} to model temporal dependencies. More recently, Transformers \cite{xu2021anomaly, tuli2022tranad, lai2023nominality} and graph neural networks (GNNs) \cite{deng2021graph, chen2021learning, zheng2023correlation, lee2026knowledge} have been widely adopted due to their effectiveness in modeling long-range temporal dependencies and inter-variable dependencies, respectively. However, due to the absence of consideration for stochasticity, these methods often struggle to distinguish true anomalies from large yet normal random fluctuations \cite{li2020anomaly, feng2024sensitivehue}.

Probabilistic approaches alleviate this limitation by incorporating likelihood-based scoring, which accounts for stochasticity through the dispersion of the predictive distribution.
In a simple formulation, some methods \cite{lin2023hybridad, feng2024time, feng2024spatial, feng2024sensitivehue} assume a parametric predictive distribution (typically Gaussian), whose parameters (e.g., mean and variance) are directly predicted by deep neural networks. Other methods \cite{zhao2020multivariate, wang2022variational, li2023prototype} model the predictive distribution by introducing window-level latent variables and training an inference network within a VAE framework. However, both approaches may be limited in capturing stochasticity, as they do not explicitly model the underlying stochastic system dynamics. Alternatively, a line of work \cite{su2019robust, li2020anomaly, dai2021sdfvae, feng2021time, li2022learning, chen2022deep, 9796836} leverages PSSMs, which separately model measurement noise and latent stochastic dynamics via emission and state-transition models, respectively, over sequantial latent variables representing system states. Existing PSSM-based MTAD methods are typically built upon RNN-based formulations, such as the variational RNN (VRNN) \cite{chung2015recurrent} and stochastic RNN (SRNN) \cite{fraccaro2016sequential}. These methods commonly condition the emission and/or transition distributions on RNN hidden states, aiming to capture long-range temporal dependencies. However, RNNs often struggle with modeling long-range temporal dependencies, while inter-variable dependencies remain insufficiently addressed. As a notable exception, Chen et al.  \cite{chen2022deep} incorporate graph convolution to aggregate cross-variable information into the RNN hidden state, which subsequently conditions the transition and emission distributions. 

In a related direction, Tian et al. \cite{tian2023variational} propose a hybrid MTAD method that combines probabilistic reconstruction with deterministic forecasting. Their probabilistic reconstruction model estimates the predictive distribution using per-time-step latent variables inferred via a Graph Transformer encoder. However, unlike PSSM-based methods, including ours, their model does not explicitly model dependencies among latent variables, limiting its ability to capture underlying stochastic system dynamics \cite{chung2015recurrent, fraccaro2016sequential}.

\section{Preliminaries}
\subsection{Problem Statement}
We consider multivariate time series with $N$ variables. A training dataset consists of observations collected over $T_\text{tr}$ time steps: $X_\text{tr} = \{x^{\text{tr}}_{1}, \dots, x^{\text{tr}}_{T_{\text{tr}}}\}$, where $x_t^{\text{tr}}\in\mathbb{R}^{N}$ represents the value of the $N$ variables at time $t$. A separate test dataset is collected over $T_{\text{te}}$ time steps: $X_\text{te} = \{x^{\text{te}}_{1}, \dots, x^{\text{te}}_{T_{\text{te}}}\}$. Since anomaly labels are scarce in practice, we assume an unsupervised setting \cite{audibert2020usad}, in which $X_\text{tr}$ contains only normal samples, while $X_{\text{te}}$ contains both normal and anomalous samples. 

Our goal is to detect anomalies in $X_{\text{te}}$. At each test time step $t$, GT-PSSM performs probabilistic reconstruction using a sliding window of observations of length $w$, denoted by $x_{t-w+1:t}^{\text{te}}\!=\!(x_{t-w+1}^{\text{te}}, \dots, x_t^{\text{te}})\!\in\!\mathbb{R}^{N\times w}$, as input. The model outputs a predictive distribution $p(\hat{x}_t^{\text{te}}|x_{t-w+1:t}^{\text{te}})$, where $\hat{x}_{t}^{\text{te}}$ denotes the reconstruction of $x_t^{\text{te}}$. The anomaly score is defined as an upper-bound surrogate of the NLL of $x_t^{\text{te}}$ under the predictive distribution. A test sample is classified as anomalous if its anomaly score exceeds a predefined threshold.

\subsection{Probabilistic State-Space Models}
PSSMs define a generative model for $x_{t-w+1:t}$ using a sequence of per-time-step latent variables $z_{t-w+1:t} = (z_{t-w+1}, \dots, z_t) \in \mathbb{R}^{d_z\times w}$, which represent the underlying system states. For each time step $l$, the latent state $z_l \in \mathbb{R}^{d_z}$ is linked to the observation $x_l$ via an emission model $p(x_l|z_l)$, while its temporal evolution is governed by a transition model $p(z_l|z_{l-1})$  \cite{murphy2012machine}. The joint distribution of $x_{t-w+1:t}$ and $z_{t-w+1:t}$ factorizes as follows:
\begin{equation}
    p(x_{t-w+1:t}, z_{t-w+1:t}) = \prod_{l=t-w+1}^{t} p(x_l|z_l)p(z_l|z_{l-1}),
    \label{eq: joint prob}
\end{equation}
where $z_{t-w}$ denotes the latent state immediately preceding the window and is typically set to $\mathbf{0}$ \cite{li2022learning}. The emission model accounts for measurement noise, while the transition model captures the inherent stochastic dynamics of the system.

To enhance modeling capacity for complex, high-dimensional multivariate time series, recent studies \cite{chung2015recurrent, fraccaro2016sequential, li2021learning} parametrize both the emission and transition distributions using deep neural networks. These deep PSSMs are typically trained within a VAE framework, where the intractable posterior over $z_{t-w+1:t}$ is approximated by an inference model $q(z_{t-w+1:t}|x_{t-w+1:t})$, also parametrized by deep neural networks. The marginal log-likelihood $\log p(x_{t-w+1:t})$, obtained by integrating out $z_{t-w+1:t}$ in Eq.\eqref{eq: joint prob}, is then optimized by maximizing the evidence lower bound (ELBO): 
\begin{align}
    \sum_{l=t-w+1}^{t}E_{q_{z_l}}[\log&\, p(x_l|z_l)]\nonumber\\& - KL(q(z_{t-w+1:t}|x_{t-w+1:t})\|p(z_{t-w+1:t})),\label{eq: elbo}
\end{align}
where  $q_{z_l} = q(z_l|x_{t-w+1:t}) = \int q(z_{t-w+1:t}|x_{t-w+1:t})\prod_{j\neq l}dz_j$ and $p(z_{t-w+1:t})=\prod_{l=t-w+1}^{t}p(z_l|z_{l-1})$.

By treating the emission model as the decoder and the inference model as the encoder, PSSMs can perform probabilistic reconstruction. Specifically, the predictive distribution of the reconstructed value of $x_t$, denoted by $\hat{x}_t$, is obtained as follows:
\begin{equation}
        p(\hat{x}_t|x_{t-w+1:t}) = \int p(\hat{x}_t|z_t)q(z_t|x_{t-w+1:t})dz_t.
        \label{eq: pred dist}
\end{equation}
PSSM-based MTAD methods typically compute  anomaly scores using either the NLL of $x_t$ under Eq.\eqref{eq: pred dist} \cite{li2020anomaly} or its upper bound derived via Jensen's inequality, $E_{q_{z_t}}[-\log p(x_t|z_t)]$~\cite{li2022learning}.

\section{Methodology}
\begin{figure*}[t]
\centering
\includegraphics[width=0.72\textwidth]{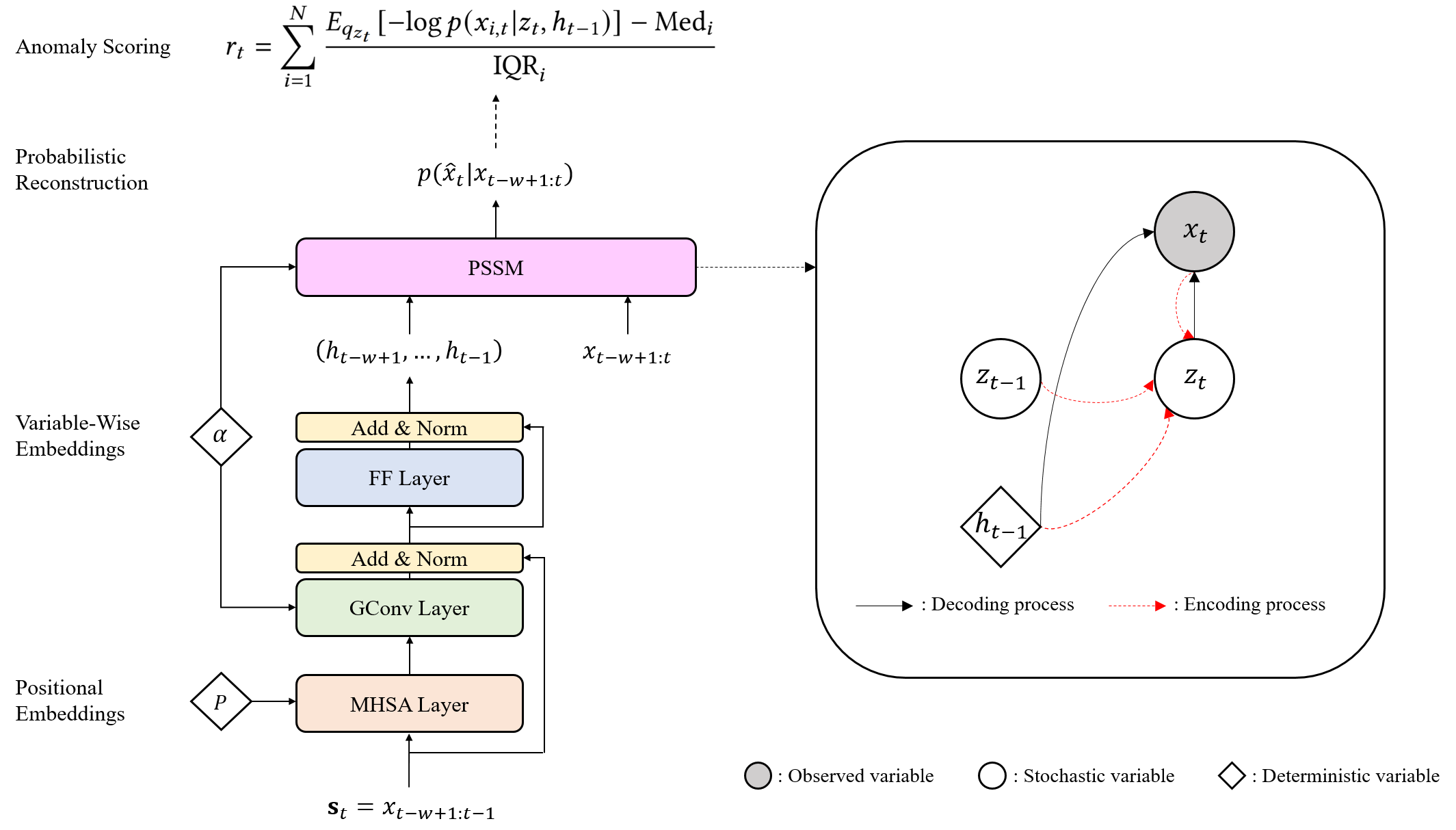}
\Description{Fully described in the text.}
\caption{Overview of GT-PSSM and a graphical representation of its encoding and decoding processes at time step $t$ for reconstructing $x_t$. The encoder and decoder correspond to the inference model in Eq.(\ref{eq: q1}) and the emission model in Eq.(\ref{eq: emission}), respectively. ``Add'' and ``Norm'' denote residual connections and layer normalization, respectively.}
    \label{fig: GT-PSSM}
\end{figure*}

An overview of GT-PSSM, together with a graphical representation of the encoding and decoding processes at time step $t$ for reconstructing $x_t$, is presented in Figure \ref{fig: GT-PSSM}. The code for GT-PSSM is available at \url{https://github.com/WonmoKoo/GT-PSSM}. 

\subsection{Stochastic Dynamics Modeling via PSSM}
We explicitly model stochasticity in multivariate time series within a PSSM framework. The generative process of $x_{t-w+1:t}$ is formulated as follows: for time step $l = t\!-\!w\!+\!1, \dots, t$,
\begin{align}
    p(x_l|z_l, h_{l-1}) &= N(W_{z}z_l + W_{h}h_{l-1} + b_x,\, \Sigma_l^x),\label{eq: emission}\\
    p(z_l|z_{l-1}, h_{l-1}) &= N(\text{MLP}([z_{l-1}, h_{l-1}]),\,
    \Sigma_l^{p_z}),\label{eq: p1}
\end{align}
where $W_{z}\in\mathbb{R}^{N \times d_z}$ and $W_{h} \in \mathbb{R}^{N \times d_h}$ are weight matrices, $b_x\in\mathbb{R}^N$ is a vector of bias,  MLP denotes a multi-layer perceptron, $[\cdot]$ denotes the concatenation operation, $\Sigma_l^x \in \mathbb{R}^{N\times N}$ and $\Sigma_l^{p_z} \in \mathbb{R}^{d_z \times d_z}$ are covariance matrices, and $z_{t-w}$ and $h_{t-w}$ are set to $\mathbf{0}$. 
All MLPs appearing in the model, including that in Eq.\eqref{eq: p1}, are distinct; their indices are omitted for notational simplicity.

To capture complex temporal and inter-variable dependencies, both the emission and transition distributions in Eqs.\eqref{eq: emission} and \eqref{eq: p1} are conditioned on $h_{l-1} \in \mathbb{R}^{d_h}$, a representation of past observations $x_{t-w+1:l-1}$ produced by a Graph Transformer (introduced in Section~\ref{subsec: graph transformer encoder}), which integrates temporal self-attention and graph convolution to model temporal and inter-variable dependencies, respectively. 

To further enhance inter-variable dependency modeling, we parameterize the emission weight matrices $W_z$ and $W_h$ in Eq.\eqref{eq: emission} using a matrix of learnable variable-wise embeddings $\alpha = (\alpha_1, \dots, \alpha_N)^{'} \in \mathbb{R}^{N \times d_e}$, motivated by \cite{bai2020adaptive, chen2021learning}, as follows:
\begin{equation}
    W_z = \alpha\delta_z, \quad W_h = \alpha\delta_h,
    \label{eq: weight matrix factorization}
\end{equation}
where $\delta_z \in \mathbb{R}^{d_e \times d_z}$, $\delta_h \in \mathbb{R}^{d_e \times d_h}$, and $d_e\!\ll\! N$. This parameterization corresponds to a low-rank  factorization of the emission weight matrices, where $\alpha$ serves as factor loadings. By sharing $\delta_z$ and $\delta_h$ across variables, variable-wise emission mappings become coupled, thereby explicitly inducing inter-variable dependency structure in the emission mean and complementing the dependency modeling via $h_{l-1}$.

For computational efficiency, we assume diagonal covariance matrices for both the emission and transition distributions, while allowing time- and variable-dependent variances to capture non-stationary measurement noise and system variability:
\begin{alignat}{2}
\Sigma_l^x &{}= \operatorname{diag}((\sigma_l^x)^{2}), \quad
& \sigma_l^x &{}= \operatorname{Softplus}(\operatorname{MLP}([z_l, h_{l-1}])), \\
\Sigma_l^{p_z} &{}= \operatorname{diag}((\sigma_l^{p_z})^2), \quad
& \sigma_l^{p_z} &{}= \operatorname{Softplus}(\operatorname{MLP}([z_{l-1}, h_{l-1}])),
\end{alignat}
where $\sigma_l^x\in\mathbb{R}^{N}$, $\sigma_l^{p_z}\in\mathbb{R}^{d_z}$, and $\operatorname{diag}(\cdot)$ constructs a diagonal matrix from its input. 

Although diagonal covariance matrices are assumed, cross-variable covariance can still be captured through the latent states. Moreover, the diagonal covariance parameterization of the emission distribution enables the application of $\beta$-NLL \cite{seitzer2022pitfalls} to the training loss in Eq.\eqref{eq:final loss}, which further contributes to stable model training.

\subsection{Dependency Modeling via Graph Transformer}
\label{subsec: graph transformer encoder}
The Graph Transformer consists of a multi-head self-attention (MHSA) layer, a graph convolution (GConv) layer, and a feed-forward (FF) layer. Residual connections \cite{he2016deep} and layer normalization \cite{ba2016layer} are applied after the GConv and FF layers. Given the input sequence $\mathbf{s}_{t} := x_{t-w+1:t-1} \in \mathbb{R}^{N\times (w-1)}$, the Graph Transformer outputs a sequence of hidden representations $\mathbf{h}_{t} = (h_{t-w+1}, \dots, h_{t-1})\in \mathbb{R}^{d_h\times(w-1)}$, where each element summarizes the observations up to the corresponding time step; for example, $h_{t-1}$ encodes information from $x_{t-w+1:t-1}$.

\subsubsection{MHSA Layer}
To effectively capture long-range temporal dependencies, we introduce an MHSA layer that performs temporal self-attention. By allowing each time step to directly attend to all preceding time steps, self-attention can capture both short- and long-range temporal dependencies, as demonstrated in prior studies \cite{chen2021learning, tang2021probabilistic}. In our MHSA layer, queries and keys are linearly projected, while an elementwise affine transformation is applied to the values to avoid mixing information across variables. Inter-variable dependency modeling is instead delegated to the subsequent GConv layer. Furthermore, learnable scalar positional embeddings \cite{wu2021scalar} are incorporated into the attention scores to encode the temporal order of the input sequence. 

The output of the MHSA layer with $M$ attention heads,  $\tilde{\textbf{h}}_t\in\mathbb{R}^{N \times (w-1)}$, is computed as follows:
\begin{gather}
        \tilde{\textbf{h}}_t =  \text{Stack}(H_1^{'}, \dots, H_M^{'})W_o, \\
        H_m = \text{Softmax}\bigg(\text{Mask}\bigg(\frac{(W_q^{m}\textbf{s}_{t})^{'}(W_k^{m}\textbf{s}_{t})}{\sqrt{d_s}} + P_m\bigg)\bigg)(\tilde{\textbf{s}}_t^{m})^{'}, \\ \tilde{\textbf{s}}_t^m = \textbf{s}_t \odot a_{m}\mathbf{1}^{'} + b_m\mathbf{1}^{'}, \quad   
        m = 1, \dots, M,
\end{gather} 
where ``Stack'' denotes the operation of stacking inputs along a new dimension; $W_o \in \mathbb{R}^{M\times 1}$, $W_q^m, W_k^m\in\mathbb{R}^{d_s\times N}$ are learnable weight matrices; $P_m \in \mathbb{R}^{(w-1)\times (w-1)}$ is a matrix of scalar positional embeddings, whose entries encode the signed temporal offsets between the query and key time steps in self-attention (see  \cite{wu2021scalar} for details); $a_m, b_m \in \mathbb{R}^N$ are vectors of weights and biases;  $\odot$ denotes the Hadamard product; and $\mathbf{1} \in \mathbb{R}^{(w-1)}$ is a vector of ones. A causal mask (denoted as ``Mask") is applied before the softmax operation so that each time step attends only to itself and preceding time steps; attention scores corresponding to future time steps are set to $-\infty$.

\subsubsection{GConv Layer} 
To explicitly model inter-variable dependencies, we employ a GConv layer, where graph nodes represent variables and edges represent their relationships. Because such relationships are typically unknown in real-world MTAD scenarios \cite{deng2021graph}, the graph structure is learned directly from data by parameterizing the adjacency matrix $A$ using the learnable variable-wise embeddings $\alpha$ (shared with Eq.\eqref{eq: weight matrix factorization}). Sharing embeddings between the emission model and the GConv layer encourages consistency in the inter-variable dependency structures captured by the two components, while simultaneously promoting similar embeddings for variables exhibiting similar behaviors.  % the components encourage them to capture characteristics of each variable, so that  variables with similar behaviors are mapped to nearby locations in the embedding space.  

The output of the GConv layer, $\bar{\textbf{h}}_t\in\mathbb{R}^{d_h \times (w-1)}$, is computed as follows:
\begin{equation}
   \bar{\textbf{h}}_t = W_g A\tilde{\textbf{h}}_t + b_g\mathbf{1}^{'}, \quad A = \text{Softmax}(\text{ReLU}(\alpha\alpha^{'})),
\end{equation}
where $W_g \in \mathbb{R}^{d_h\times N}$ is a learnable weight matrix, $b_g \in \mathbb{R}^{d_h}$ is a vector of biases, and $\text{ReLU}$ denotes the rectified linear unit. The resulting adjacency matrix defines a fully connected weighted graph over variables.

\subsubsection{FF Layer} 
Finally, we apply a position-wise feed-forward network (i.e., an MLP applied independently at each time step) to obtain the final representations $\mathbf{h}_t$ as follows:
\begin{equation}
     \mathbf{h}_{t} = (h_{t-w+1}, \dots, h_{t-1}) =  \text{MLP}(\bar{\textbf{h}}_t).
\label{eq: FFN}
\end{equation}

\subsection{Model Training}
To train GT-PSSM, we derive an ELBO for $\text{log}\,p(x_{t-w+1:t})$ by approximating the posterior distribution over $z_{t-w+1:t}$ using the inference model $q$:
\begin{equation}
    q(z_{t-w+1:t}|x_{t-w+1:t}) = \prod_{l=t-w+1}^{t}q(z_l|z_{l-1}, h_{l-1}, x_{l}),
    \label{eq: q1}
\end{equation}
where $q(z_l|z_{l-1}, h_{l-1}, x_{l}) =  N(\text{MLP}([z_{l-1}, h_{l-1}, x_l]), \, \text{diag}((\sigma_l^{q_z})^2)))$ and $\sigma_l^{q_z} = \text{Softplus}(\text{MLP}([z_{l-1}, h_{l-1}, x_l])) \in \mathbb{R}^{d_z}$. The ELBO is computed in the same manner as in Eq.\eqref{eq: elbo}.

However, in our experiments, minimizing the standard negative ELBO using  gradient-based optimization often failed to accurately estimate the mean of $p(x_l|z_l, h_{l-1})$, consistent with the findings of \cite{seitzer2022pitfalls}. This issue arises because, under Eq.\eqref{eq: emission}, the reconstruction term in the negative ELBO corresponds to the expected Gaussian NLL, whose gradients with respect to the mean are scaled by the variance. As a result, observations with large predicted variances may contribute negligibly to mean estimation. To mitigate this issue, we adopt $\beta$-NLL \cite{seitzer2022pitfalls} and reweight the KL-divergence terms accordingly. The final training loss is computed by approximating expectations using a single Monte Carlo sample from $q$ as follows:
\begin{align}
\label{eq:final loss}
    \mathcal{L}=  \sum_{l=t-w+1}^{t}&\bigg(\sum_{i=1}^{N}-\lfloor(\sigma^x_{i, l})^{2\beta}\rfloor\text{log}\,p(x_{i, l}|z_l, h_{l-1})\nonumber \\&+\lambda_l KL(q(z_l|z_{l-1}, h_{l-1}, x_l)||p(z_l|z_{l-1}, h_{l-1}))\bigg),  
\end{align}
where $\sigma_{i,l}^{x}$ is the $i$-th element of $\sigma_l^x$, $x_{i, l}$ is the observed value of variable $i$ at time step $l$, and $\lambda_l = \frac{1}{N}\sum_{i=1}^{N}\lfloor(\sigma_{i, l}^{x})^{2\beta}\rfloor$, with $\lfloor \cdot \rfloor$ denoting the stop-gradient operation and $\beta$ representing a hyperparameter. The model parameters are learned by minimizing $\mathcal{L}$.

\subsection{Anomaly Scoring}
\label{subsec: anomaly scoring}
For anomaly detection, GT-PSSM performs probabilistic reconstruction. To compute the anomaly score at time step $t$, we use an upper bound on the NLL of $x_t$ under $p(\hat{x}_t|x_{t-w+1:t})$:
\begin{equation}
E_{q_{z_t}}[-\log p(x_t|z_t, h_{t-1})] = \sum_{i=1}^{N} E_{q_{z_t}}[-\log p(x_{i,t}|z_t, h_{t-1})],
\end{equation}
where $q_{z_t} = q(z_t|x_{t-w+1:t}) = \int q(z_t|z_{t-1}, h_{t-1}, x_t)q_{z_{t-1}}dz_{t-1}$. In practice, the dynamics of different variables may vary substantially, resulting in heterogeneous scales of their anomaly score contributions. Following \cite{deng2021graph} and \cite{feng2024sensitivehue}, we therefore normalize the contribution of each variable using the median and interquartile range (IQR) of the corresponding quantity computed over all time steps in a validation dataset. The final anomaly score at time step $t$ is computed as follows: 
\begin{equation}
    r_t = \sum_{i=1}^N \frac{E_{q_{z_t}}[-\text{log}\,p(x_{i, t}|z_t, h_{t-1})] - \text{Med}_i}{\text{IQR}_i},
    \label{eq: anomaly score}
\end{equation}
where $\text{Med}_{i}$ and $\text{IQR}_{i}$ denote the median and IQR for variable $i$, respectively. In our experiments, $E_{q_{z_t}}$ is approximated using a sufficiently large number of Monte Carlo samples ($L \gg\!1$) to adequately account for reconstruction uncertainty.

\begin{table*}[ht]
\caption{Anomaly detection performances of GT-PSSM and the baselines in terms of \textbf{F1} and  $\textbf{F1}_{\text{PA}}$. The baselines are ordered chronologically, with earlier methods listed at the top and more recent ones at the bottom. Bold and underlined values indicate the best and second-best performance, respectively. %AnoTrans denotes Anomaly Transformer \citep{xu2021anomaly}.
}\label{table: performance comparison}
\begin{center}
\resizebox{\textwidth}{!}{%
\begin{tabular}{c|cc|cc|cc|cc}
\hline\hline
\multirow{2}{*}{\textbf{Method}}  &  \multicolumn{2}{c|}{\textbf{WADI}} & \multicolumn{2}{c|}{\textbf{PSM}} & \multicolumn{2}{c|}{\textbf{SMD}} & \multicolumn{2}{c}{\textbf{SWaT}} \\
\cline{2-9}
&\textbf{F1}&$\textbf{F1}_{\textbf{\text{PA}}}$&\textbf{F1}&$\textbf{F1}_{\textbf{\text{PA}}}$&\textbf{F1}&$\textbf{F1}_{\textbf{\text{PA}}}$&\textbf{F1}&$\textbf{F1}_{\textbf{\text{PA}}}$\\
\hline
MAD-GAN&0.3415 (0.0446)&0.4601 (0.0752)&0.4539 (0.0306)&0.7140 (0.0908) &0.2027 (0.0539)& 0.3823 (0.1052) &0.6126 (0.1137)&0.7046 (0.1081)\\
USAD &0.5217 (0.0247)&0.6435 (0.0299)&0.4861 (0.0046)&0.7263 (0.0134) &0.6264 (0.0117)& 0.9203 (0.0085)&0.7517 (0.0016)& 0.8218 (0.0000) \\
MTAD-GAT&0.5302 (0.0084)&0.8059 (0.0511)&0.5806 (0.0073)&0.8158 (0.0222)& 0.5259 (0.0302)& 0.9978 (0.0001)&0.7722 (0.0029)&0.8766 (0.0076)\\
GDN&0.5306 (0.0272)&0.7564 (0.0323)&0.5488 (0.0368)&0.7806 (0.0159) &\underline{0.6807} (0.0935)& 0.9953 (0.0100)&0.8186 (0.0143)&0.8797 (0.0021)\\
GTA&0.5165 (0.0079)&\underline{0.8333} (0.0047)&0.5398 (0.0110)& 0.9670 (0.0039) &0.3793 (0.0104)& \underline{0.9996} (0.0002)&0.7768 (0.0085)&0.8954 (0.0062) \\
NSIBF&0.4894 (0.0314)&0.6527 (0.0397)& 0.5921 (0.0241)&0.9116 (0.0104)&0.5800 (0.1222)& 0.9964 (0.0027)&0.7917 (0.0054)&0.9150 (0.0054)\\
AnoTrans &0.0705 (0.0084)&0.7527 (0.0195)&0.4009 (0.0170)&\textbf{0.9815} (0.0028)&0.0257 (0.0048)&0.9982 (0.0007)&0.1544 (0.0608)&0.9026 (0.0106) \\
DVGCRN &0.5391 (0.0109)&0.7090 (0.0177)&0.4456 (0.0219)&0.7838 (0.0721)&0.3162 (0.0840)&0.9242 (0.0869)&0.7580 (0.1363)&0.8310 (0.0558)\\
TranAD&0.5901 (0.0104)&0.8008 (0.0175)&0.5282 (0.0050)& 0.7795 (0.0007) &0.6604 (0.0214)& 0.9983 (0.0004)&0.8108 (0.0001)&0.8692 (0.0032) \\
CST-GL&0.5697 (0.0097)& 0.7287 (0.0395)& 0.4346 (0.0001)&\underline{0.9814} (0.0043)&0.5929 (0.0613)& \textbf{0.9999} (0.0002)&0.7696 (0.0008)&0.8875 (0.0073) \\
NPSR&0.5676 (0.0165)&0.6609 (0.0165)&\underline{0.6335} (0.0201)&0.7588 (0.0239)&0.6003 (0.0427)& 0.9321 (0.0163)&0.7735 (0.0016)& 0.8265 (0.0009) \\
SensitiveHUE&\underline{0.6749} (0.0075)&0.8076 (0.0130)& 0.5244 (0.0038)&0.8327 (0.0143)&0.6514 (0.0211)& 0.9988 (0.0002)&\textbf{0.8530} (0.0053)&\textbf{0.9318} (0.0068)\\
\hline
GT-PSSM&\textbf{0.7343} (0.0095)&\textbf{0.9320} (0.0186)&\textbf{0.6631} (0.0076)& 0.8484 (0.0362)&\textbf{0.7375} (0.0229) & 0.9985 (0.0000)&\underline{0.8466} (0.0049)&\underline{0.9168} (0.0057)\\
\hline\hline
\end{tabular}}
\end{center}
\end{table*}

\section{Experiments}
\subsection{Dataset Description}
To validate GT-PSSM, we used four benchmark datasets collected from complex real-world systems: \textbf{1) WADI} (WAter DIstribution) \cite{ahmed2017wadi} was collected from a water distribution system with 127 sensors and actuators over 14 days of normal operation and 2 days of operation under cyber attacks; \textbf{2) PSM} (Pooled Server Metrics) \cite{abdulaal2021practical} is a multivariate time series dataset with 25 variables collected from an application server cluster at eBay, where the training and test dataset contain 13 and 8 weeks of data, respectively; \textbf{3) SMD} (Server Machine Dataset) \cite{su2019robust} was collected from a large internet company and comprises multiple time series of 38 variables from 28 independent server machines, each spanning 10 days, with the first half used for training and the second half for testing; and \textbf{4) SWaT} (Secure Water Treatment) \cite{mathur2016swat} was collected from a water treatment test-bed system with 51 sensors and actuators over 11 days, where the final approximately 5 days constitute the test dataset and contain cyber-attack intervals. Dataset statistics are summarized in Table \ref{table: dataset statistics} in Appendix \ref{appendix: dataset statistics}.

\subsection{Baselines}
We compared GT-PSSM with 12 baselines. \textbf{1) Deterministic approaches:} MAD-GAN \cite{li2019mad}, USAD \cite{audibert2020usad}, MTAD-GAT \cite{zhao2020multivariate}\footnote{The original paper employed a VAE for probabilistic reconstruction within a hybrid framework. However, due to the absence of an official implementation, we used a publicly available implementation in which the VAE is replaced with a deterministic autoencoder.}, GDN \cite{deng2021graph}, GTA \cite{chen2021learning}, Anomaly Transformer (AnoTrans) \cite{xu2021anomaly}, TranAD \cite{tuli2022tranad},  CST-GL~\cite{zheng2023correlation}, and NPSR \cite{lai2023nominality}; and \textbf{2) Probabilistic approaches:} NSIBF \cite{feng2021time}, DVGCRN
\cite{chen2022deep}, and SensitiveHUE \cite{feng2024sensitivehue}. 

Implementation details of GT-PSSM are provided in Appendix \ref{appendix: implementation details}. The results of the baselines were obtained using either the official code released by the original authors or publicly available implementations, whose sources are listed in Appendix~\ref{appendix: source baseline}. Particularly, in the official implementations of GDN \cite{deng2021graph} and SensitiveHUE \cite{feng2024sensitivehue}, $\mathrm{Med}_i$ and $\mathrm{IQR}_i$ used in the anomaly scores are computed over all test time steps, requiring prior access to the entire test dataset and thus limiting applicability in online settings. Therefore, we modified the implementations to compute these quantities using a validation dataset instead.

\subsection{Evaluation Metrics}
We evaluate anomaly detection performance using the F1 score, the harmonic mean of precision and recall. Many previous MTAD studies adopt the point-adjusted F1 score \cite{xu2018unsupervised}, in which all time steps within a consecutive anomalous segment are counted as correctly detected if at least one time step in the segment is identified as anomalous. However, this adjustment has been shown to overestimate performance \cite{kim2022towards, doshi2022reward, lai2023nominality}. Therefore, we report both the original (unadjusted) F1 score ($\mathbf{F1}$) and the point-adjusted F1 score ($\mathbf{F1}_{\text{PA}}$), while adopting  $\mathbf{F1}$ as the primary evaluation metric. Following common practice \cite{lai2023nominality, liu2024mtad, feng2024sensitivehue, shimillas2026low}, we determine the anomaly detection threshold via grid search and report the best score for each metric.

\subsection{Comparison with Baselines}
\label{subsec: comparison with baselines}
Table \ref{table: performance comparison} summarizes the results averaged over 10 repeated experiments (standard deviations in parentheses). In terms of $\mathbf{F1}$, GT-PSSM achieved the best performance on three out of the four datasets (WADI, PSM, and SMD) and remained highly competitive on SWaT. Specifically, compared to the best-performing baseline on each dataset, GT-PSSM improved $\mathbf{F1}$ from 0.6749 to \textbf{0.7343 (+5.94\%)} on WADI, from 0.6335 to \textbf{0.6631 (+2.96\%)} on PSM, and from 0.6807 to \textbf{0.7375 (+5.68\%)} on SMD. On SWaT, GT-PSSM achieved the second-best $\mathbf{F1}$, only slightly below the highest score  (0.8466 vs. 0.8530).

In terms of $\mathbf{F1}_{\text{PA}}$, GT-PSSM achieved the best performance on WADI and remained competitive on SMD and SWaT. However,  its $\mathbf{F1}_{\text{PA}}$ on PSM was relatively lower. We believe this may be attributable to conservative anomaly annotations in PSM: time steps adjacent to long labeled anomalous segments often exhibit similar anomalous patterns but are annotated as normal. Consequently, improving sensitivity to such patterns may increase both true positives and false positives. Under $\mathbf{F1}_{\text{PA}}$, additional true positives within the same anomalous segment receive no additional credit, whereas false positives are fully penalized, resulting in a lower score even when anomalies are captured more faithfully. 

\begin{figure*}[!ht]
  \centering
  \begin{subfigure}[t]{\linewidth}
    \centering
    \includegraphics[width=0.72\textwidth]{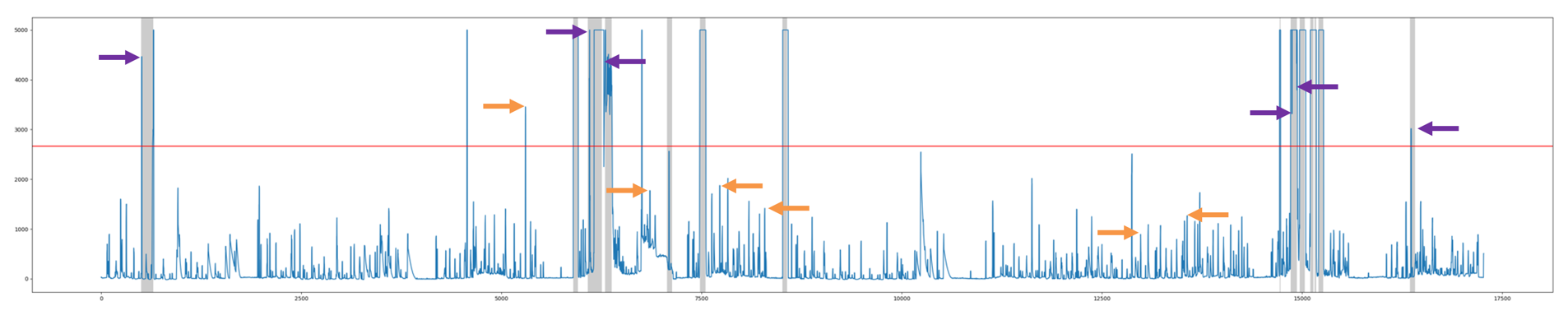}
    \caption{GT-PSSM}
    \label{subfig: qualitative gtpssm}
    \Description{Fully described in the text.}
  \end{subfigure}\\
  \begin{subfigure}[t]{\linewidth}
    \centering
    \includegraphics[width=0.72\textwidth]{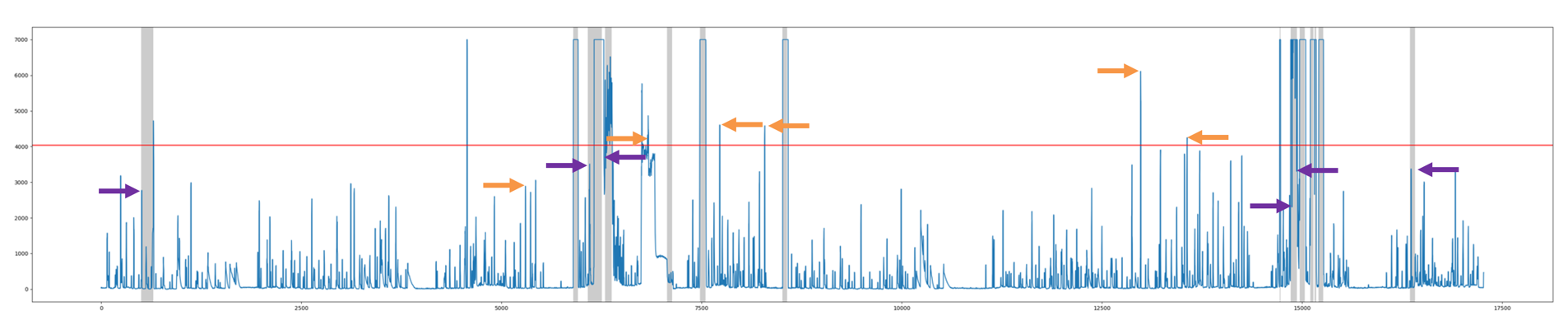}
    \caption{w/o Unc.}
    \label{subfig: qualitative wouncertainty}
    \Description{Fully described in the text.}
  \end{subfigure}
  \caption{Visualization of anomaly scores on WADI produced by GT-PSSM and its variant (``w/o Unc.''). The gray shaded regions indicate ground-truth anomalous segments, while the red horizontal lines denote the anomaly detection threshold. Purple and orange arrows indicate anomalous and normal data points, respectively, for which GT-PSSM and the variant produce contrasting detection outcomes.}
  \label{fig: qualitative}
\end{figure*}

Consistent with this explanation, Table~\ref{table: performance comparison} shows an inverse trend between $\mathbf{F1}$ and $\mathbf{F1}_{\text{PA}}$ among methods that outperform GT-PSSM in $\mathbf{F1}_{\text{PA}}$ (AnoTrans, CST-GL, GTA, and NSIBF). Notably, the top two methods in terms of $\mathbf{F1}_{\text{PA}}$ rank lowest and second-lowest in terms of $\mathbf{F1}$, highlighting a trade-off between the two metrics.

\begin{table}[t]
\caption{Results of the ablation study evaluated using \textbf{F1}. Bold and underlined values indicate the best and second-best performance, respectively.}
\label{table: ablation study}
\begin{center}
\resizebox{0.48\textwidth}{!}{%
\begin{tabular}{c|c|c|c|c}
\hline\hline
\textbf{Method}&\textbf{WADI}  & \textbf{PSM} & \textbf{SMD} & \textbf{SWaT} 
\\
\hline
\multicolumn{1}{l|}{GT-PSSM} &\textbf{0.7343} (0.0095) &\textbf{0.6631} (0.0076) &\textbf{0.7375} (0.0229)&\textbf{0.8466} (0.0049)\\
\multicolumn{1}{l|}{w/o Unc.} &0.7055 (0.0209) & 0.6450 (0.0066)&0.7115 (0.0315)&0.8435 (0.0036)\\
\multicolumn{1}{l|}{ w/o Trans.} &\underline{0.7193} (0.0109)& \underline{0.6577} (0.0029)& 0.7197 (0.0137)&\underline{0.8448} (0.0025)\\
\multicolumn{1}{l|}{ w/o MHSA} &0.6891 (0.0301)& 0.6464 (0.0096) & 0.7156 (0.0160)&0.8306 (0.0076)\\
\multicolumn{1}{l|}{ w/o GConv} &0.7127 (0.0083)&0.6377 (0.0136)&0.7154 (0.0269)&0.8405 (0.0078)\\
\multicolumn{1}{l|}{ w/o LRMF} &0.7026 (0.0088)&0.6054 (0.0089)&0.6954 (0.0121)& 0.8314 (0.0171)\\
%\multicolumn{1}{l|}{ w/o GConv \& LRMF} &0.6803 (0.0173)& 0.5326 (0.0100) & 0.6538 (0.0520)&\\
\multicolumn{1}{l|}{ w/o $\beta$-NLL} &0.6857 (0.0076) &0.6384 (0.0211)&\underline{0.7344} (0.0107)&0.8128 (0.0036)\\
\hline\hline
\end{tabular}}
\end{center}
\end{table}

\subsection{Ablation Study}
\label{sec: ablation study}
We conducted an ablation study to evaluate the contributions of the main components of GT-PSSM, using $\mathbf{F1}$ as the evaluation metric. The results are summarized in Table~\ref{table: ablation study}.  

\textbf{Uncertainty-aware anomaly scoring.} First, to examine the importance of incorporating reconstruction uncertainty into anomaly scoring, we evaluated a variant (``w/o Unc.'') that computes the anomaly score at each time step using the squared error between the observation and the mean reconstruction, with variable-wise $\text{Med}$-$\text{IQR}$ normalization applied as described in Section \ref{subsec: anomaly scoring}. GT-PSSM outperformed this variant on WADI, PSM, and SMD, improving \textbf{F1} by \textbf{+2.88\%}, \textbf{+1.81\%}, and \textbf{+2.60\%}, respectively, while achieving comparable performance on SWaT. These results suggest that point-wise reconstruction or forecasting errors alone are often insufficient for anomaly detection in stochastic environments. 

\textbf{State-transition modeling.} Second, we evaluated a variant without the state-transition model (``w/o Trans.''), in which the direct connections between stochastic latent states in Eqs.\eqref{eq: p1} and \eqref{eq: q1} were removed. Removing this component led to decreases in $\mathbf{F1}$ on WADI, PSM, and SMD, while yielding comparable performance on SWaT. These results suggest that explicitly modeling temporal dependencies among latent states improves the representation of stochastic system dynamics.

\textbf{Temporal and inter-variable dependency modeling.} Third, we examined three additional variants: (i) replacing the Graph Transformer with a graph-convolutional RNN (``w/o MHSA'') to assess the contribution of temporal self-attention to modeling long-range temporal dependencies, (ii) removing the GConv layer (``w/o GConv'') to assess the contribution of graph convolution to inter-variable dependency modeling, and (iii) disabling the low-rank weight matrix factorization (LRMF) in Eq.\eqref{eq: weight matrix factorization} (``w/o LRMF'') to assess the contribution of low-rank structural bias in variable-wise emissions to inter-variable dependency modeling. All three variants showed significant performance degradation, highlighting the advantages of temporal self-attention for capturing long-range temporal dependencies and of explicit mechanisms for modeling inter-variable dependencies. 

Interestingly, removing the LRMF resulted in a larger performance drop than removing the GConv layer. We do not interpret this as evidence that the LRMF is inherently more important than the GConv layer for inter-variable dependency modeling. Rather, the effect of removing the LRMF likely reflects not only the loss of a structured low-rank inductive bias on the emission weights, but also the loss of alignment between the variable-specific emission mappings and the inter-variable dependency structure learned by the GConv layer. This likely explains the greater degradation observed in the ``w/o LRMF'' variant. We note that isolating the effect of the low-rank inductive bias alone is not possible in GT-PSSM, since removing the LRMF also affects the GConv layer through parameter sharing.

\textbf{Incorporation of $\beta$-NLL into the training loss.} Finally, we evaluated a variant without $\beta$-NLL (``w/o $\beta$-NLL''), in which GT-PSSM was trained by optimizing the standard ELBO, corresponding to $\beta=0$. This variant showed noticeable performance degradation on WADI, PSM, and SWaT, while achieving comparable performance on SMD. These results suggest that $\beta$-NLL stabilizes the learning of the emission mean and thereby contributes to enhancing Graph Transformer-based dependency learning in GT-PSSM.

Overall, each component contributes meaningfully, and their integration yields the best performance. 

\subsection{Qualitative Analysis}
\subsubsection{Effectiveness of Uncertainty-Aware Anomaly Scoring}
We visualize the WADI results from the first ablation setting in Section~\ref{sec: ablation study} to qualitatively demonstrate the advantage of uncertainty-aware anomaly scoring over point-wise-error-based scoring. Figures \ref{subfig: qualitative gtpssm} and ~\ref{subfig: qualitative wouncertainty} show anomaly scores during the test period produced by GT-PSSM and its ``w/o Unc.'' variant, respectively. In each panel, the red horizontal line indicates the anomaly detection threshold that maximizes \textbf{F1}, where observations exceeding the threshold are classified as anomalous. Gray shaded regions denote ground-truth anomalous segments. Purple and orange arrows highlight anomalous and normal data points, respectively, for which GT-PSSM and the variant produced contrasting detection outcomes. As highlighted by the purple arrows, GT-PSSM successfully detected true anomalies missed by the variant. Conversely, as indicated by the orange arrows, GT-PSSM correctly classified these points as normal in all but one case, whereas the variant falsely labeled them as anomalies.

Closer inspection shows that the orange-marked time steps correspond to abrupt increases in water flow, which induce pronounced stochastic fluctuations. By explicitly modeling such stochasticity, uncertainty-aware (likelihood-based) anomaly scoring in GT-PSSM avoids spurious detections during benign variability, thereby leading to more accurate anomaly detection.

\begin{figure}
    \centering
    \includegraphics[width=0.75\linewidth]{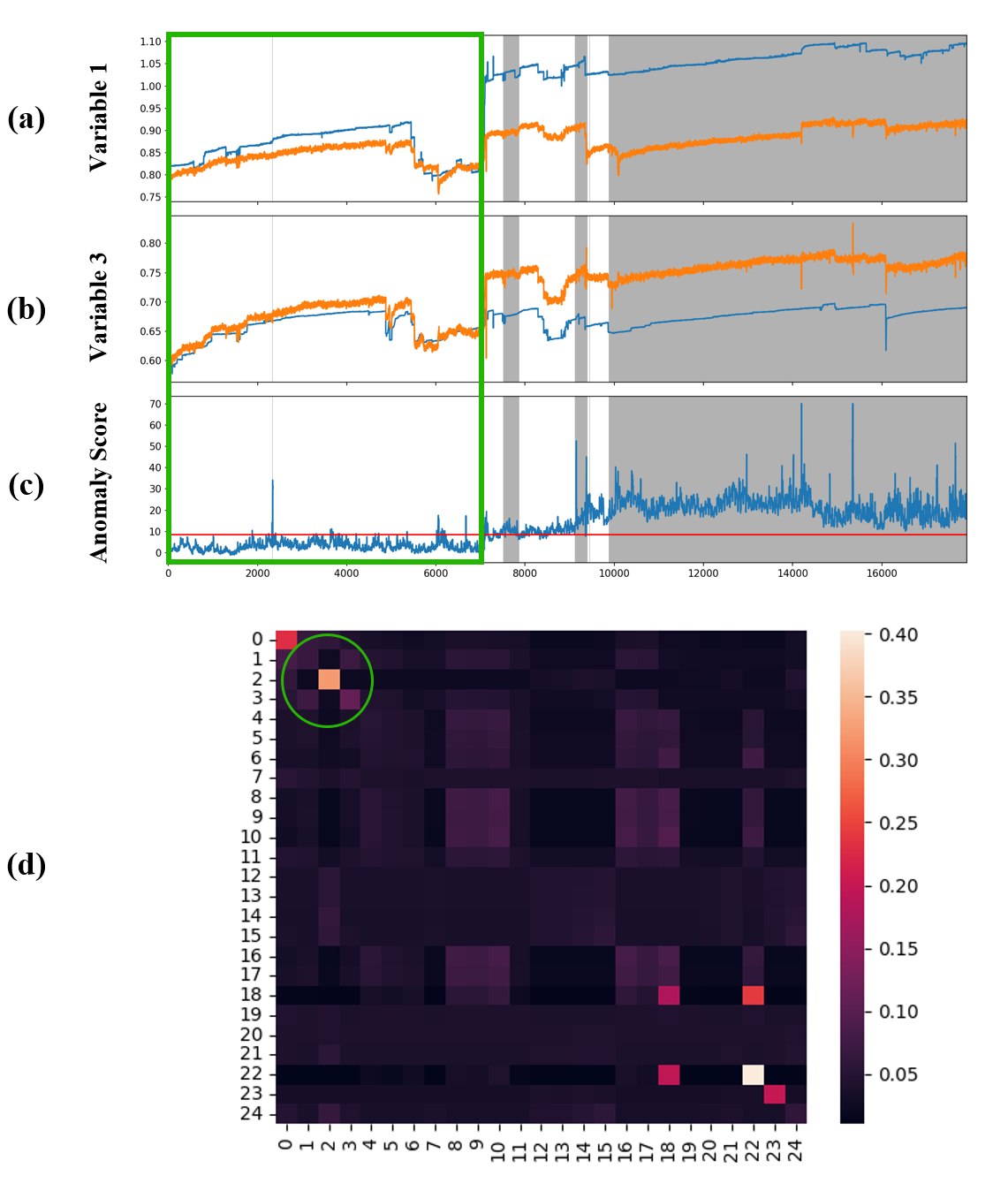}
    \caption{(a)--(b): Observations (blue line) and mean reconstructions (orange line) for Variables 1 and 3 over a selected portion of the PSM test period. The normal period is highlighted by the green box; (c): Anomaly scores over the same period, together with the threshold that maximizes $\mathbf{F1}$ (red horizontal line); (d): Heat map of the adjacency matrix of the GConv layer (rows: target nodes, columns: neighboring nodes), where lighter colors indicate stronger dependencies.}
    \label{fig: qualitative inter-variable dependencies}
    \Description{Fully described in the text.}
\end{figure}

\subsubsection{Importance of Inter-Variable Dependency Modeling}
To illustrate the importance of modeling inter-variable dependencies in MTAD---particularly in cases where individual variables do not exhibit clear anomalous signals, but anomalies instead emerge through changes in cross-variable dependency structure---we present results from GT-PSSM for a scenario in which each variable independently appears normal, yet anomalies arise from abnormal relationships between variables.

Figure \ref{fig: qualitative inter-variable dependencies} shows test results on PSM near a long anomalous segment. Figures~\ref{fig: qualitative inter-variable dependencies}a and~\ref{fig: qualitative inter-variable dependencies}b display the observations (blue line) and mean reconstructions (orange line) for the two variables with the largest contributions to the anomaly score during the anomalous segment. Figure \ref{fig: qualitative inter-variable dependencies}c shows the anomaly scores, together with the threshold that maximizes $\mathbf{F1}$ (red horizontal line). 
The time axis and anomalous segments (gray shaded regions) are identical across Figures \ref{fig: qualitative inter-variable dependencies}a--\ref{fig: qualitative inter-variable dependencies}c. Figure \ref{fig: qualitative inter-variable dependencies}d presents a heat map of the adjacency matrix learned by the GConv layer from the training data and used during graph convolution in reconstruction. Rows correspond to target nodes (variables), columns correspond to neighboring nodes, and lighter colors indicate stronger dependencies.

As shown in Figure~\ref{fig: qualitative inter-variable dependencies}a and~\ref{fig: qualitative inter-variable dependencies}b, during the normal period (green box), the observations of \textbf{Variable 1} and \textbf{Variable 3} (blue lines) remain similar in magnitude over time. GT-PSSM captures this relationship, which is also reflected in the learned adjacency matrix (green circle in Figure \ref{fig: qualitative inter-variable dependencies}d). In contrast, during the long anomalous segment, the two variables diverge substantially: \textbf{Variable 1} maintains a high level and continues to increase, whereas \textbf{Variable 3} remains within a value range similar to that observed during the normal period. 
Since GT-PSSM reconstructs each variable based on learned inter-variable dependencies, this disruption of the learned relationship leads to large deviations between the reconstructions and observations, resulting in elevated anomaly scores during the anomalous period, as shown in Figure \ref{fig: qualitative inter-variable dependencies}c.

\subsection{Training Dynamics Analysis}
Recent studies \cite{feng2024sensitivehue, sunmultivariate} have identified an over-generalization issue in reconstruction-based MTAD methods: anomalous samples are often reconstructed nearly as well as normal ones, thereby degrading anomaly detection performance. This phenomenon is attributed to the model’s tendency to learn an identity mapping that simply replicates the input, rather than capturing the underlying normal patterns (i.e., temporal and inter-variable dependencies) \cite{feng2024sensitivehue}. Consequently, improved training data fit can paradoxically lead to worse anomaly detection performance \cite{you2022unified}.

To examine whether this issue also arises in GT-PSSM, we monitored the average marginal log-likelihood on training sliding windows, along with  $\mathbf{F1}$ and $\mathbf{F1}_{\text{PA}}$ on the test dataset, every five epochs during training. As the marginal log-likelihood is intractable, we used the ELBO as a practical surrogate. We emphasize that this protocol---evaluating test performance during training---was adopted solely for this analysis; in all other experiments, anomaly detection performance on the test dataset was not assessed during training.

\begin{figure}
\centering
\includegraphics[width=0.7\linewidth]{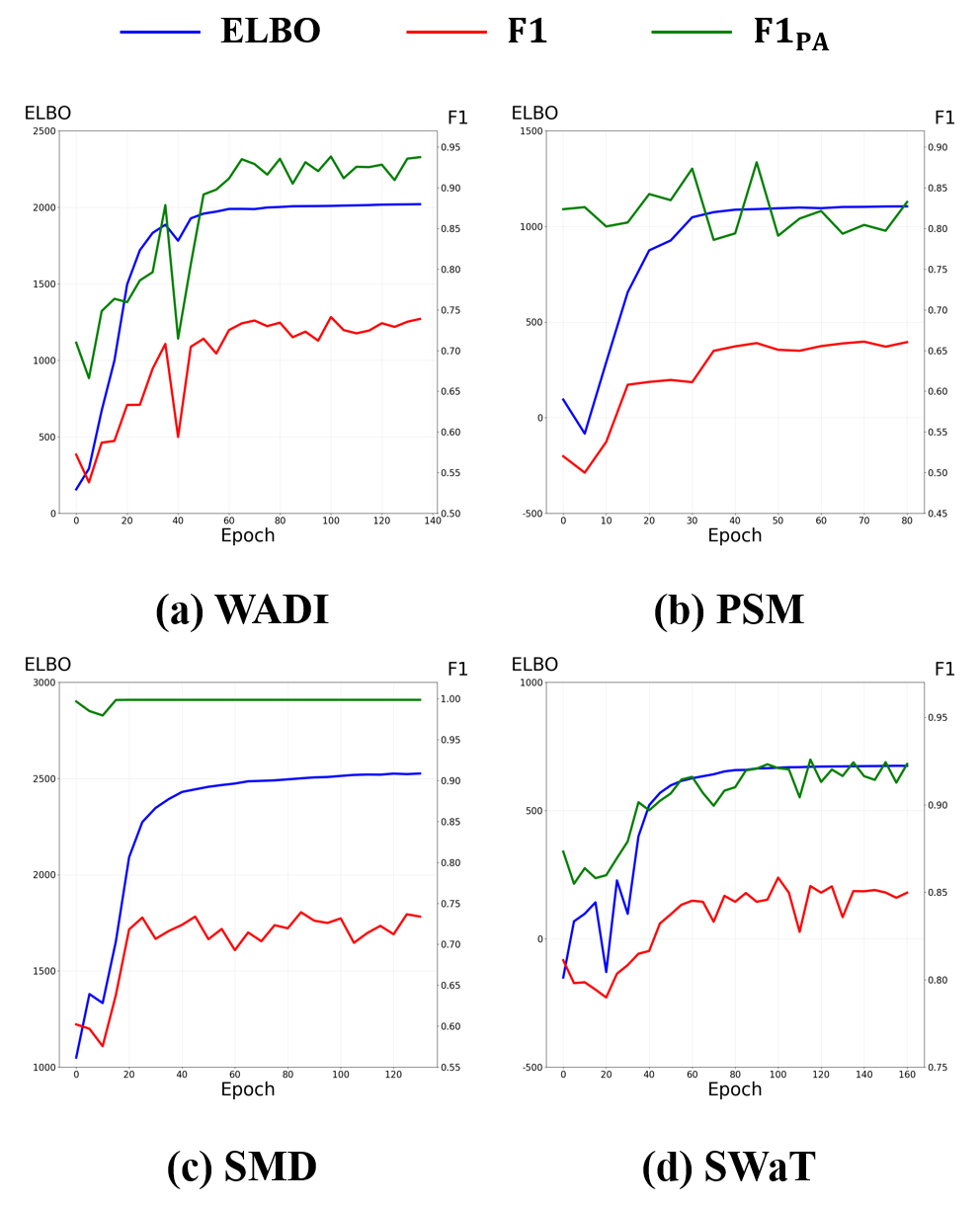}
\caption{Average ELBO on training windows, together with $\mathbf{F1}$ and $\mathbf{F1}_{\text{PA}}$ on the test dataset, evaluated every five epochs during training for each dataset.}
\label{fig: training dynamics analysis}
\Description{Fully described in the text.}
\end{figure}

As shown in Figure \ref{fig: training dynamics analysis}, across all four datasets, the ELBO, $\mathbf{F1}$, and $\mathbf{F1}_{\text{PA}}$ generally increase as training progresses, except for $\mathbf{F1}_{\text{PA}}$ on PSM. These trends indicate that better training data fit is associated with improved anomaly detection performance in GT-PSSM, suggesting that over-generalization is unlikely to be a major concern in our model. 

\subsection{Hyperparameter Sensitivity Analysis}
\begin{figure}
\centering
\includegraphics[width=0.7\linewidth]{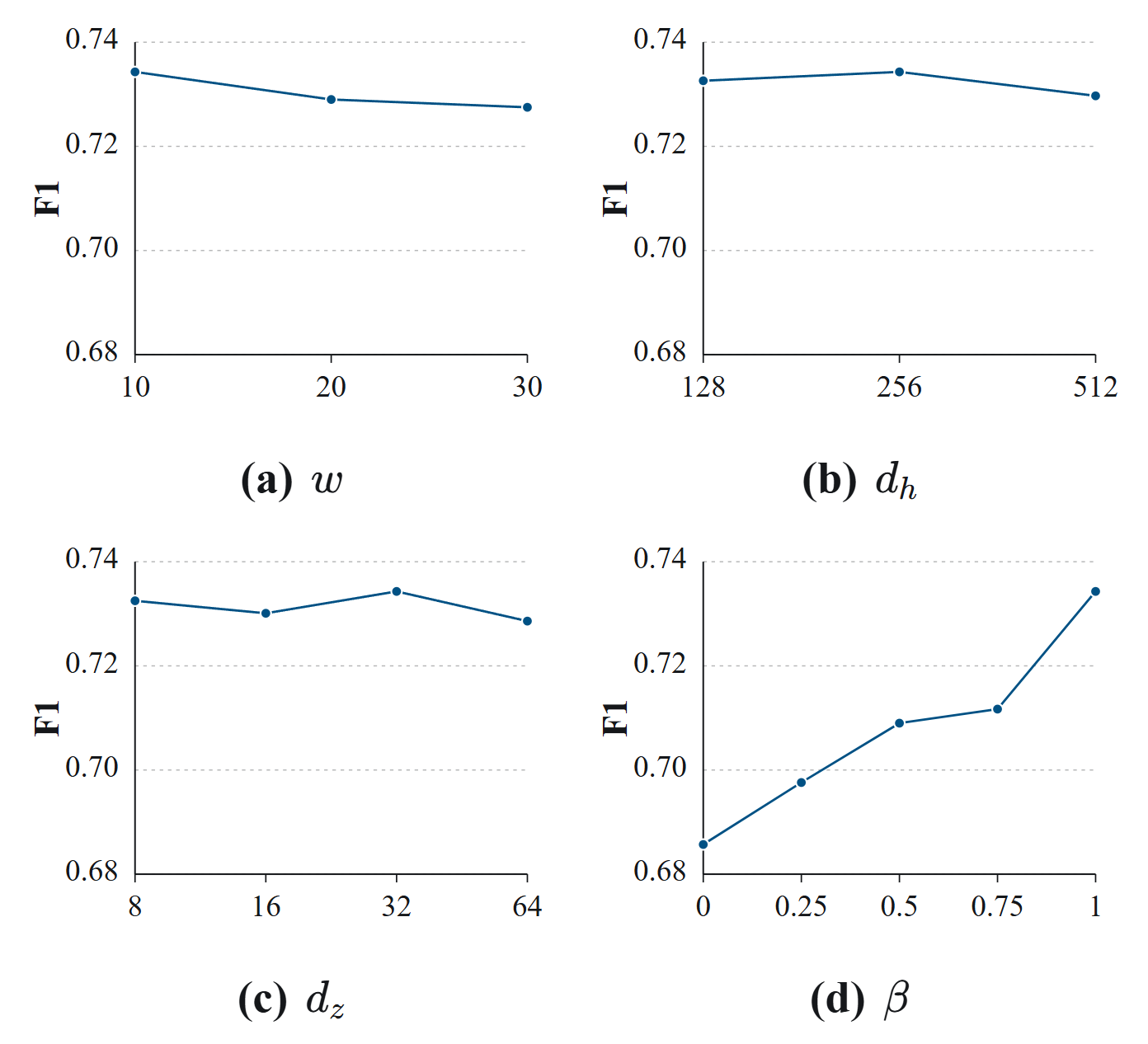}
\caption{The results of the sensitivity analysis on WADI.}
\label{fig: hyper sen}
\Description{Fully described in the text.}
\end{figure}

To investigate the effects of the key hyperparameters in GT-PSSM, namely $w$, $d_h$, $d_z$, and $\beta$, we evaluated anomaly detection performance in terms of \textbf{F1} while varying their values. As shown in Figure \ref{fig: hyper sen}, GT-PSSM remained relatively stable across different hyperparameter settings, exhibiting no substantial performance variation with respect to $w$, $d_h$, and $d_z$.

By contrast, performance was more sensitive to the choice of $\beta$ in the training loss (Eq.\eqref{eq:final loss}). Specifically, \textbf{F1} consistently increased as $\beta$ increased. Because a larger $\beta$ reduces the influence of the predicted variance on mean learning while preserving uncertainty modeling, it can stabilize mean estimation. Since GT-PSSM captures inter-variable dependencies mainly through the mean of $p(x_l|z_l, h_{l-1})$, this may explain the observed trend. Based on these observations, we fixed $\beta$ at 1.0 across all datasets to avoid dataset-specific hyperparameter tuning.

\section{Conclusion}
We introduced GT-PSSM, a novel MTAD framework that tightly integrates a Graph Transformer within a probabilistic state-space model to jointly capture stochastic dynamics, long-range temporal dependencies, and inter-variable dependency structures in complex multivariate time series.  
Unlike prior PSSM-based MTAD methods, GT-PSSM overcomes key limitations in dependency modeling through a principled integration strategy: parameterizing the transition and emission distributions through Graph Transformer representations, aligning variable-wise emission means with the learned inter-variable dependency structure, and leveraging $\beta$-NLL to stabilize emission mean learning and, consequently, Graph Transformer training. By unifying probabilistic state-space modeling with Graph Transformer-based dependency learning, GT-PSSM preserves principled uncertainty modeling while enhancing the representation of complex temporal and inter-variable dependencies. Extensive experiments on four real-world benchmark datasets demonstrated that GT-PSSM achieves state-of-the-art anomaly detection performance and consistently outperforms strong deterministic and probabilistic baselines. 
%We proposed GT-PSSM, a novel PSSM-based MTAD method that jointly models  stochasticity, long-range temporal dependencies, and inter-variable dependency structures in complex multivariate time series. Through probabilistic reconstruction and uncertainty-aware anomaly scoring, GT-PSSM achieved accurate anomaly detection. By embedding a Graph Transformer within a probabilistic state-space framework, GT-PSSM overcomes key limitations of prior PSSM-based MTAD methods in capturing long-range temporal and inter-variable dependencies while retaining principled stochasticity modeling. 
%Beyond anomaly detection, our framework highlights the potential of tightly integrating structured dependency learning with probabilistic temporal modeling in multivariate time series analysis. 
Beyond anomaly detection, GT-PSSM suggests a broader direction for multivariate time series learning: integrating structured inductive biases for dependency learning with principled probabilistic representations of stochastic dynamics.
An important direction for future work is to extend the learned graph in GT-PSSM from associative dependencies to causal structures, enabling interpretable root-cause analysis of detected anomalies.
% Despite its strong anomaly detection performance, GT-PSSM may be less effective under concept drift, where the underlying data distribution changes over time. As future work, we plan to develop online concept drift adaptation strategies for GT-PSSM to maintain reliable detection in non-stationary environments.

%\begin{acks}
%  ...
%\end{acks}

\appendix
\section{Dataset Statistics}
\label{appendix: dataset statistics}
The statistics of the datasets used in our experiments are summarized in Table \ref{table: dataset statistics}. For WADI, we followed the experimental settings of \cite{deng2021graph}: the first 21,600 training samples were discarded due to system instability during that period, a subset of 112 variables was selected, and the data were downsampled at every 10 time steps. For SMD, although the original dataset comprises multivariate time series from 28 independent server machines, we used only \texttt{machine-1-1}. Following prior studies \cite{tuli2022tranad, zheng2023correlation}, which selected subsets of machines to exclude those exhibiting severe concept drift \cite{zheng2023correlation} or containing overly trivial anomalies \cite{tuli2022tranad}, we selected \texttt{machine-1-1} because it was included in the subsets used by both studies. For SWaT, following \cite{deng2021graph}, the first 21,600 training samples were discarded and the data were downsampled at every 10 time steps. Variables AIT201 and P201 were excluded because their distributions in the training data differed noticeably from those in the test data under normal operating conditions \cite{perales2020madics}, following \cite{lai2023nominality}. For all datasets, Min-Max normalization was applied, and the validation ratio was set to 0.2.
%The statistics of the datasets used in our experiments are summarized in Table \ref{table: dataset statistics}. For WADI, we follow the experimental settings of \cite{deng2021graph}. Specifically, we discarded the first 21,600 training samples as the underlying system was unstable during that period, and selected a subset of 112 variables. The data were then downsampled every 10 time steps. For SMD, even though the original dataset consists of multivariate time series from 28 independent server machines, we used only the trace named \texttt{machine-1-1}, a non-trivial series without severe concept drift issues \cite{tuli2022tranad, zheng2023correlation}. For all datasets, we applied Min-Max normalization and set the validation ratio to 0.2.
\begin{table}[ht]
    \caption{Statistics of the datasets used in the experiments: number of variables, number of training time steps, number of test time steps, and the ratio of anomalies in the test data.}
        \vspace{-0.12 in}
    \begin{center}
    \resizebox{0.35\textwidth}{!}{%
    \begin{tabular}{c|c|c|c|c}
    \hline
    \hline
    \textbf{Dataset}&\textbf{\#Variables}&\textbf{\#Train}&\textbf{\#Test}&\textbf{\%Anomalies}\\
    \hline
    WADI& 112 &  118,800& 17,280& 6  \\
    PSM& 25 &  132,481& 87,841&28\\
    SMD& 38 &  28,479 & 28,479& 9\\
    SWaT & 49 & 47,520 & 44,991&12\\
    \hline
    \hline
    \end{tabular}}
    \end{center}
    \label{table: dataset statistics}
        \vspace{-0.12 in}
\end{table}

\section{Implementation Details}
\label{appendix: implementation details}
All experiments were conducted using a TITAN V GPU. For all datasets, GT-PSSM was trained using the Adam optimizer~ \cite{kingma2014adam} with a learning rate of $10^{-3}$ for up to 500 epochs. Early stopping was applied with a patience of 20 epochs. The batch size was set to 128 for WADI, PSM, and SWaT, and 32 for SMD.  We used two-layer MLPs, with ReLU activation for the MLP in Eq.\eqref{eq: FFN} and Tanh activation for the remaining MLPs. For WADI, PSM, and SWaT, all MLPs had 256 and 128 hidden units in the first and second hidden layers, respectively, whereas for SMD, the corresponding numbers were 128 and 64. The remaining hyperparameter settings of GT-PSSM for each dataset are summarized in Table \ref{table: hyperparameter settings}.
\begin{table}[ht]
    \caption{Detailed hyperparameter settings for GT-PSSM.}
    \vspace{-0.12 in}
    \begin{center}
    \resizebox{0.32\textwidth}{!}{%
    \begin{tabular}{c|c|c|c|c|c|c|c|c}
    \hline
    \hline
    \textbf{Dataset}&$w$&$d_h$&$d_z$&$d_e$&$d_s$&$M$&$\beta$&$L$\\
    \hline
    WADI&10&256&32&32&32&8&1.0&200\\
    PSM&20&256&16&6&32&8&1.0&200\\
    SMD&30&128&8&6&16&8&1.0&200\\
    SWaT&10&256&32&8&32&8&1.0&200\\
    \hline
    \hline
    \end{tabular}}
    \end{center}
    \label{table: hyperparameter settings}
    \vspace{-0.12 in}
\end{table}

\section{Sources of Baseline Implementations}
\label{appendix: source baseline}
\begin{itemize}
\raggedright
    \item MAD-GAN: \url{https://github.com/LiDan456/MAD-GANs}
    \item USAD: \url{https://github.com/manigalati/usad}
    \item MTAD-GAT: \url{https://github.com/ML4ITS/mtad-gat-pytorch}
    \item GDN: \url{https://github.com/d-ailin/GDN}
    \item GTA: \url{https://github.com/zackchen-lb/GTA}
    \item NSIBF: \url{https://github.com/cfeng783/NSIBF}
    \item Anomaly Transformer: \url{https://github.com/thuml/Anomaly-Transformer}
    \item DVGCRN: 
    \url{https://github.com/BoChenGroup/DVGCRN}
    \item TranAD: \url{https://github.com/imperial-qore/TranAD}
    \item CST-GL: \url{https://github.com/huankoh/CST-GL}
    \item NPSR: \url{https://github.com/andrewlai61616/NPSR}
    \item SensitiveHUE: \url{https://github.com/yuesuoqingqiu/SensitiveHUE}
\end{itemize}

\section{Evaluation on the Full SMD}
\label{app:full smd}
Although our main experiments on SMD were conducted using only \texttt{machine-1-1}, we additionally evaluated GT-PSSM on the full SMD dataset, which comprises 28 machines, to examine its robustness. Table \ref{table:smd_full} reports the average \textbf{F1} over all machines and compares GT-PSSM with several representative strong baselines. Specifically, we included GDN, which achieved the second-highest \textbf{F1} on SMD in our main experiments; NPSR, which achieved the second-highest \textbf{F1} on PSM; and SensitiveHUE, which achieved the highest \textbf{F1} on SWaT and the second-highest \textbf{F1} on WADI. As shown in Table \ref{table:smd_full}, GT-PSSM achieved the highest average \textbf{F1}, demonstrating its robustness on the full SMD dataset.

\begin{table}[ht]
\caption{Anomaly detection performance of GT-PSSM and the selected baselines on the full SMD dataset comprising all 28 machines, evaluated using \textbf{F1}. We report the mean and standard deviation of the machine-averaged \textbf{F1} scores across 10 repeated experiments. The value in bold indicates the best performance, and the underlined value denotes the second-best performance.}
\label{table:smd_full}
\vspace{-0.12 in}
\begin{center}
\resizebox{0.44\textwidth}{!}{%
\begin{tabular}{ccc|c}
\hline\hline
\multicolumn{4}{c}{\textbf{SMD (Full)}}\\
\hline
\textbf{GDN} &\textbf{ NPSR} & \textbf{SensitiveHUE} & \textbf{GT-PSSM}\\\hline
0.5209 (0.0064) & \underline{0.5233} (0.0114) & 0.4250 (0.0084) & \textbf{0.5272} (0.0113)\\
\hline\hline
\end{tabular}}
\end{center}
\vspace{-0.12 in}
\end{table}

\section{Computational Cost Analysis}
Table \ref{table:computational cost} reports the training time per epoch (TPE, in seconds), time per forward pass (TPF, in seconds), and memory usage (MiB) for GT-PSSM and the baselines selected in Appendix \ref{app:full smd}. TPF and memory usage were measured at the batch level during inference using a unified batch size of 128 for WADI and 32 for SMD across all methods. GT-PSSM incurred relatively higher computational costs. Nevertheless, with a TPF of less than 0.1 seconds per batch on both datasets, GT-PSSM remains feasible for online inference in practical deployment scenarios.

\begin{table}[ht]
\caption{Computational cost comparison on WADI and SMD. The values in bold indicate the highest computational costs, while the underlined values denote the second-highest costs. }
\vspace{-0.12 in}
\label{table:computational cost}
\begin{center}
\resizebox{0.44\textwidth}{!}{%
\begin{tabular}{c|c|c|c|c}
\hline\hline
\multirow{2}{*}{\textbf{Dataset}} & \multirow{2}{*}{\textbf{Method}} & \textbf{TPE} & \textbf{TPF} & \textbf{Memory Usage} \\
& & \textbf{(seconds)} & \textbf{(seconds)} & \textbf{(MiB)} \\
\hline
\multirow{4}{*}{WADI}&GDN&25.82&0.017&\textbf{2169}\\
& NPSR &\textbf{180.27}&\underline{0.034}&1466\\
& SensitiveHUE &11.25&0.003&134\\
\cline{2-5}
& GT-PSSM &\underline{52.58}&\textbf{0.076}&\underline{1832}\\
\hline
\multirow{4}{*}{SMD}&GDN&3.31&0.001&72\\
& NPSR &\underline{42.16}&\underline{0.004}&\textbf{593}\\
& SensitiveHUE &8.73&0.003&112\\
\cline{2-5}
& GT-PSSM &\textbf{66.94}&\textbf{0.049}&\underline{591}\\\hline

\hline\hline
\end{tabular}}
\end{center}
\vspace{-0.12 in}
\end{table}

\section{Acknowledgments}
This research was supported by the National Research Foundation of Korea (NRF) grant funded by the Korea government
(MSIT) (2023R1A2C2005453, RS-2023-00218913).

\newpage
\section*{GenAI Usage Disclosure}
The authors used ChatGPT and Claude exclusively for English language proofreading.  
%to improve the clarity of the manuscript.

\bibliographystyle{ACM-Reference-Format}
\balance
\bibliography{references}

\end{document}